\documentclass{article}

\usepackage[preprint]{ewrl_2026}

\usepackage[utf8]{inputenc} 
\usepackage[T1]{fontenc}    
\usepackage{hyperref}       
\usepackage{url}            
\usepackage{booktabs}       
\usepackage{amsfonts}       
\usepackage{nicefrac}       
\usepackage{microtype}      
\usepackage{xcolor}         

\usepackage{amssymb}            
\usepackage{mathtools}          
\usepackage{mathrsfs}           

\usepackage{mycommands}
\usepackage{bbm}
\usepackage{booktabs}
\usepackage{multirow}
\usepackage{color}
\definecolor{lightgray}{gray}{0.90}
\newcommand\greybox[1]{%
  \vskip\baselineskip%
  \par\noindent\colorbox{lightgray}{%
    \begin{minipage}{\textwidth}#1\end{minipage}%
  }%
  \vskip\baselineskip%
}

\title{Locally-Guided Actor-Critic: Training a Goal-conditioned Actor with a Subgoal-aware Critic}

\author{ 
    Olivier Serris \\ \texttt{serris@isir.upmc.fr} \\
    ISIR\\
    Sorbonne Université, CNRS
    \And
    St\'{e}phane Doncieux \\\texttt{Stephane.Doncieux@isir.upmc.fr}\\
    ISIR\\
    Sorbonne Université, CNRS
    \And
    Olivier Sigaud \\
    \texttt{Olivier.Sigaud@isir.upmc.fr}\\
    ISIR\\
    Sorbonne Université, CNRS
}

\begin{document}

\maketitle

\begin{abstract}
Goal-conditioned reinforcement learning struggles with long horizons when rewards are sparse. While a planner can provide subgoals to guide a low-level policy, its use at test time may introduce practical subgoal management difficulties. An alternative paradigm utilizes a high-level planner to assist learning,  while the policy remains conditioned only on the final goal, enabling planner-free deployment. Among these methods, Reinforcement Learning with Imagined Subgoals (RIS) introduces a regularization term that encourages the policy to take the same actions for the final goal as it does for an intermediate goal. This regularization, however, may lead to goal-chaining issues when intermediate goals are low-dimensional. Potential-based reward shaping (PBRS) translates plans into an additional reward while ensuring that the optimal policy remains unchanged. Yet, it can generate deceptive rewards in terminal states. We study these failure cases and first propose an alternative reward shaping method (RS) that removes these deceptive rewards at the expense of theoretical guarantees of PBRS.  Similar to this RS variant, we then propose another method named Locally-Guided Actor Critic (LG-AC) that rewards the agent for reaching intermediate goals. Unlike RS, where intermediate rewards are implicit in the shaping signal, we explicitly condition a value estimator on the full sequence of intermediate goals but represent the value function as a sum of subgoal-conditioned value functions, enabling dense hindsight relabeling. We evaluate all these methods in tasks with challenging goal-chaining requirements and empirically highlight specific cases in which either action regularization or reward shaping yield low performance, while LG-AC achieves the best overall performance across tasks.
\end{abstract}

\section{Introduction}
In Reinforcement Learning (RL), an agent typically learns to master a task by maximizing a reward signal over a sequence of actions. However, standard RL methods often struggle when horizons are long and rewards are sparse. Goal-Conditioned (GC) policies \citep{schaul2015universal} are no exception, frequently failing to reach distant goals without additional guidance. To address this, hierarchical reinforcement learning (HRL) approaches decompose the task into a hierarchy of subtasks. While HRL encompasses a vast range of approaches, most follow three primary families: the option framework \citep{doina_options_smdp}, where a high-level policy iteratively selects low-level sub-policies to execute, the feudal framework \citep{dayan_feudal_92} where a high-level policy outputs intermediate goals that a low-level policy must reach iteratively, or graph-based planning \citep{savinov_semi_parametric} where the low-level policy is similarly conditioned on intermediate goals, but these goals are generated via search over a graph of previously visited states. While effective for long-horizon tasks, these approaches typically require the high-level planner to operate during inference to guide the low-level policy which can lead to increased computation cost or subgoal management difficulties \citep{serris2025talegoalsleveragingsequentiality}. An orthogonal line of research only uses the high-level as a guidance for a low-level policy during learning, while the low-level policy operates on its own, conditioned solely on the final goal. With this paradigm, the high-level planner can utilize privileged information, such as ground-truth simulation states or global maps, that is only available in simulation.

Within this category of methods, \ris \citep{chane-sane_goal-conditioned_2021} proposes to constrain the policy to output the same action for a far-away final goal as it would for a goal placed midway between its current location and the final goal. 
However, \cite{chenu_stil} highlights a general chaining issue that can arise in similar contexts: when intermediate goals are low-dimensional and thus do not fully specify the state, a goal can be reached in a state incompatible with the achievement of the rest of the trajectory.  
To avoid these problems, other methods use potential-based reward shaping (PBRS) \citep{ng_policy_inv} to convert high-level plans into rewards. This method preserves the optimal policy of the original MDP, but respecting PBRS constraints can create deceptive rewards in terminal states \citep{kudenko_improv_eff_pbrs}. Consequently, the agent is either incentivized to terminate in any previously visited terminal state or discouraged from reaching any terminal state, including goals.

In this paper, we evaluate an alternative method based on reward shaping (RS) that rewards reaching each intermediate goal while removing deceptive rewards in terminal states, at the expense of PBRS's theoretical guarantees. Additionally, we propose a second method inspired by the RS approach but with a key architectural difference. In RS, the agent receives implicit rewards for reaching subgoals through a shaped reward signal. In another approach named Locally-Guided Actor Critic (\ourmethodABR), this structure is made explicit by conditioning a value estimator on the complete sequence of intermediate goals provided by a high-level planner. This structure enables hindsight relabeling for intermediate goals. To maintain a fixed input size regardless of the length of the plan, \ourmethodABR decomposes the value into a sum of per-goal components. This decomposition leads to an estimator conditioned on two distinct goal types simultaneously: the final goal, corresponding to the goal targeted by the policy, and an intermediate goal that provides an auxiliary reward. The actor is optimized using this estimator to reach all intermediate goals, which constitutes the plan for a given final goal.

We present an experimental study comparing methods using the \ris loss, \pbrs, \rs, and \ourmethodABR. All methods share the same SAC+HER backbone and a handcrafted planner to ensure that each success or failure can be directly attributed to the low-level policy. We evaluate these methods in five goal-conditioned environments where the agent must learn to reach any goal from any state. Our results show that \ris performs well in all environments except \hopper and \walker, where goal chaining issues arise. The \pbrs variant struggles when the environment contains terminal states, while the \rs variant performs well except in \antmazeu. Our \ourmethodABR method achieves the best overall mean and interquartile mean (IQM) across all environments.

\section{Preliminaries}
\label{sec:preliminary}

In this work, we study goal-conditioned reinforcement learning (GCRL) agents that leverage hindsight relabeling and formalize the task of completing sequences of goals within the framework of a Partially Observable Markov Decision Process. This section provides the necessary background for these elements.

\subsection{Markov Decision Process}
\label{sec:MDP}

We consider the standard RL framework, where an agent learns to make decisions through interaction with an environment, modeled as a Markov Decision Process (MDP) defined by the tuple $\{S,A,R,P,\rho,\mathcal{T},\gamma \}$, where $S$ is the state space, $A$ is the action space. $R:S\times A \times S \rightarrow  \mathbb{R}$ is the reward function and $P:S\times A\times S\rightarrow [0,1]$ is the probability of transitioning to a new state given a state and action. $\mathcal{T}$ corresponds to the set of terminal states. The discount factor $\gamma \in [0,1)$ balances immediate and future rewards. 
At the beginning of an episode, an initial state is sampled from $\rho$. At each discrete time step $t$, the agent chooses an action $a_t\in A$, moves to a new state $s_{t+1}$ and receives a reward $r_t$. The objective is to find a policy that maximizes the discounted cumulative reward: 
\begin{equation*}
        \operatorname*{\mathbb{E}}_{\substack{
        a_t\sim\pi(\cdot\mid s_t) \\
        s_0\sim \rho
        }}
        \left[
        \sum_{t=0}^\infty
        \gamma^t R(s_t,a_t,s_{t+1})
        \prod_{i=1}^{t} \mathbbm{1}{[s_i\notin \mathcal{T}]}
        \right].
\end{equation*}

\paragraph{Partially Observable Markov Decision Process:}
When the agent cannot directly observe the state $s$, 
Partially Observable Markov Decision Processes (POMDPs) extend MDPs by adding an observation space $\Omega$ and an observation function $O:S \rightarrow \Omega$, where the agent selects actions based solely on observations $o= O(s)$. 

\subsection{Goal-conditioned Reinforcement Learning}
\label{sec:gcrl}

Goal-conditioned Reinforcement Learning (GCRL) extends the standard MDP framework to settings where an agent must learn to reach multiple objectives. We formalize this as a GC-MDP defined by the tuple $(S_{gc},A,R,P,\rho,\mathcal{T},\gamma)$, where the state space $S_{gc} = S\times G$ is the Cartesian product of the original state space $S$ and a goal space $G$. Within this framework, the agent aims to learn a universal policy $\pi(a|s,g)$ that generalizes across different goals \citep{schaul2015universal}. To relate states to goals, we define a mapping function $\phi: S \rightarrow G$ that projects a state into the goal space. In this work, $G$ is a low-dimensional representation (e.g., the $(x,y)$ coordinates of an agent). 
We consider a sparse reward function for goal reaching, defined as $R_{gc}(s',g)= \mathbbm{1}[s'\in S_g]$, where $S_g$ represents the set of states that satisfy goal $g$. Since this reward only depends on the achieved state $s'$ and the goal $g$, we simplify notation and write $R_{gc}(s',g)$ rather than $R_{gc}(s, a, s', g)$.

\paragraph{Actor-Critic Methods:} A popular approach for solving GCRL problems is to use actor-critic methods. In these methods, a critic learns a goal-conditioned action-value function $Q(s, a, g)$ measuring the expected return of taking action $a$ in state $s$ while pursuing goal $g$ and then following the current policy. The actor $\pi(s, g)$ is updated to maximize the value provided by the critic. Off-policy algorithms are especially advantageous in GCRL because they let the agent learn from past experiences across different goals. During training, both the actor and the critic are updated using transitions $(s_t, a_t, s_{t+1}, g, r_t)$ sampled from a replay buffer. In our work, we use Soft Actor-Critic (SAC) \citep{haarnoja_soft_2018} which adds an entropy maximization term to promote exploration.

\paragraph{Hindsight experience replay:} GCRL with sparse rewards is challenging: as long as an agent does not reach a given goal, it collects no reward, hindering learning. Hindsight Experience Replay (HER) \citep{Andrychowicz_her} addresses this difficulty by replacing a failed goal in a trajectory with a goal the agent actually reached. In off-policy algorithms, each transition can be relabeled independently as $(s_t, a_t, s_{t+1}, g) \rightarrow (s_t, a_t, s_{t+1}, \phi(s_{t+k}))$, where $k>t$ is a future time step in the same trajectory. These relabeled transitions are then used in standard temporal-difference updates.

\section{Related Work}

Our approach addresses goal-conditioned Markov decision processes (GC-MDPs) by leveraging high-level plans composed of low-dimensional goals during training. We combine an asymmetric actor-critic formulation with value decomposition to distill planner-based rewards into a low-level GC-policy. We review methods that leverage intermediate goals or plans to learn flat goal-conditioned policies.

\paragraph{Policy Regularization via Subgoals:} In RIS \citep{chane-sane_goal-conditioned_2021}, a high-level policy generates intermediate goals and the low-level policy is constrained to produce similar actions when targeting a distant goal as when targeting an intermediate goal along the same path. This is implemented by minimizing a KL divergence between the action distributions. The intuition is to re-use knowledge of how to reach nearby goals to make progress toward distant ones. PIG \citep{kim2023imitating} follows a similar idea but replaces the KL divergence with a quadratic penalty between actions. Due to this similarity, we only evaluate RIS in our experiments. A similar loss to RIS is also used in SAW \citep{zhou2025flattening} in an offline RL framework. Instead of learning a high-level policy to propose intermediate goals, it directly samples intermediate states from relevant trajectories in the offline dataset, with each subgoal weighted differently depending on its relevance to reaching the final goal. In PIG and in our study, the goals are low-dimensional, often corresponding to the location of the agent. In contrast, RIS and SAW directly use states as goals but rewards are typically defined with respect to a low-dimensional goal such as the agent location. As a result, it remains to be shown whether these methods truly learn to reach exact states, or whether they effectively only target the rewarded element of the goal-state. In the latter case, goal chaining issues can arise because a valid action to reach an intermediate goal may differ significantly from a valid action to reach a distant goal along the same trajectory.  In contrast, our agent is optimized to reach all the goals of a given plan, thus encouraging actions that are compatible with both immediate and future low-dimensional subgoals. 

\paragraph{Plan-Based Reward Shaping:}
An alternative way to convert plans into learning signals for flat policies is reward shaping, a technique that augments the original reward function with additional rewards to guide learning. A particular instance is potential-based reward shaping (PBRS), in which the additional reward is defined through a difference of potential over states. This formulation comes with a theoretical guarantee that preserves the optimal policy of the initial MDP \citep{ng_policy_inv}.
In \cite{Okudo_subgoal_based_rs}, the authors design a PBRS reward defined over a sequence of goals, providing positive rewards when the agent reduces the remaining length of the sequence and penalties when it increases. Their PBRS reward is only defined for a single sequence of goals. We extend this idea to multi-goal RL in our baselines. In our method, we only provide positive rewards when the agent reaches intermediate goals. However, we use a critic explicitly conditioned on intermediate goals to allow relabeling for these goals.
In \cite{lo_gsp}, the authors propose to solve a high-level options SMDP and then use the critic of this high-level SMDP to define a potential-based reward to guide a flat policy. Their approach, however, is primarily applied in tabular and discrete-action RL, whereas we target problems with continuous states and actions.

\section{Methods}
We propose a method for solving goal-reaching tasks. During training, an expert planner guides policy learning by providing a sequence of intermediate goals towards any final goal. At deployment, the learned policy operates without access to the planner and is conditioned only on the final goal. We first formalize a problem where an agent receives rewards for intermediate goals via an asymmetric actor-critic method: the actor only sees the final goal, while the critic has access to the full goal sequence. We then show that the critic’s value estimate for a state can be written as a sum of discounted returns associated with each intermediate goal. Finally, we present an algorithm called Locally-Guided Actor-Critic (LG-AC) that jointly trains the actor and the critic using this additive structure.

\subsection{Intermediate goals as privileged information}
\label{sec:goal_as_priv_info}
In this section, we formalize an optimization problem where the agent is rewarded for reaching each intermediate goal of a given plan. By formulating the problem as a POMDP, we treat these intermediate goals as part of the hidden state, making them unobservable to the actor, ensuring it can be deployed without a planner. Additionally, to facilitate learning, we use an asymmetric actor-critic formulation \citep{pinto_asym_act_crit}, in which the critic has access to the intermediate goals as privileged information.

We define a \seqgoalmdp (\sgmdp) as a specific POMDP in which the agent is rewarded for intermediate goals that are part of the hidden state. We note it
$$
\text{\sgmdpmath}=
\{S_{seq},A,O,\Omega,G,\rsum,P,\rho,\mathcal{T},\gamma \},
$$
where each component inherits from the POMDP definition, Section \ref{sec:MDP}. The key distinction lies in the state space $S_{seq}$, which augments the underlying environment state with the current plan. At the start of the episode the agent starts in state $s_0$ with goal $\gpi \in G$ that stays fixed during the whole episode. The planner computes a sequence of goals 
$$\gseq{_0}=(g_1,g_2,\ldots,\gpi) = plan(s_0,g_\pi),$$
which defines a path from the initial state to the final goal. We use bold notation for sequences. 

The hidden state of \sgmdpmath is defined as $(s_t,\boldsymbol{g}_t,\gpi)\in S_{seq} = S\times G^+\times G$, which corresponds to augmenting the observation with a sequence of intermediate goals. $G^+$ denotes the set of non-empty finite sequences of goals from $G$, $G^+ = \bigcup_{i \ge 1} G^i$, where $G^i$ is the set of sequences of length $i$ from $G$.
After taking action $a_t$, the agent transitions to a new hidden state $(s_{t+1},\gseq_{t+1},\gpi)$, where $s_{t+1}\sim P(.|s_t,a_t)$. 
The actor only observes $(s_t,\gpi)=O(s_t,\boldsymbol{g}_t,\gpi) $, the environment state and the final goal. 
The intermediate goal sequence  $\gseq_{t+1}$ is updated by removing any goal that has been achieved according to
\begin{equation}
\label{eq:goal_transition}
\gseq_{t+1} =
\gseq_t \setminus \{g\in \gseq_t | s_{t+1} \in S_g\} 
\end{equation}
where $S_g$ corresponds to the set of states that satisfy goal $g$.
The reward function is defined as 
\[\rsum(s_{t+1},\gseq_t) = \sum\limits_{\gr\in \gseq_t}\rgc(s_{t+1},\gr).
\]
Thus each goal contributes to rewards only until reached. The objective is to maximize the expected discounted return:
\begin{equation}
\label{eq:vsg}
        V_{sg}(s,\gseq,\gpi) =
        \operatorname*{\mathbb{E}}_{\substack{
        a_t\sim\pi(\cdot\mid s_t,\gpi) \\
        }}
        \left[
        \sum_{t=0}^\infty
        \gamma^t\rsum(s_{t+1},\gseq_t)
        \prod_{i=1}^{t} \mathbbm{1}{[s_i\notin \mathcal{T}]}
        \middle| \substack{s_0=s\\\gseq_0=\gseq}
        \right],
\end{equation}
where $\mathcal{T}$ denotes the set of terminal states and $\gseq_t$ is defined in \eqref{eq:goal_transition}.

Compared to the classical GC-MDP framework, the \sgmdp framework provides intermediate rewards along a planned path, resulting in denser learning signals. However, this formulation requires the value function to be conditioned on $\gseq_t$, a sequence whose length varies as goals are achieved. This dependency introduces two related challenges: the critic must generalize over a combinatorial number of possible goal subsets, and it must operate on variable-size inputs, which significantly increases architectural complexity.

\subsection{Value Decomposition over individual goals}

To address the two challenges outlined in Section~\ref{sec:goal_as_priv_info},
we propose decomposing the critic into a sum of goal-specific values. Specifically, we aim to replace a critic function that takes a sequence of intermediate goals as input with a parameterized estimator that takes a single intermediate goal as input and returns the corresponding goal-related return. 
We can then sum the outputs of this estimator over the full sequence of goals to recover the original return estimate. As established in prior work on reward decomposition \citep{juozapaitis_expl_rew_dec}, such a summation remains valid in off-policy settings, and is compatible with temporal difference learning, ensuring the global value function is accurately preserved.
This decomposition allows us to learn a parameterized estimator with fixed-size inputs, enabling the use of simple multi-layer neural networks, while still handling sequences of arbitrary length. 

Because rewards are additive and goals can be reached independently of one another, the total discounted return for reaching all goals of the sequence admits an exact decomposition into a sum of per-goal contributions.
Each of these contributions is defined as \(\ourv(s,\gr,\gpi)\), the expected discounted return associated with the rewards for achieving \(\gr\) under a policy conditioned on \(\gpi\): 
\begin{equation}
\label{eq:vtransit}
\ourv(s,\gr,\gpi)
\triangleq\\
\mathbb{E}_{a\sim\pi(\cdot\mid s,\gpi)}
\left[
\sum_{t=0}^{\infty}
\gamma^t
\rgc(s_{t+1},\gr)
\prod_{i=1}^{t}
\mathbbm{1}{[s_i\notin (S_{\gr}\cup\mathcal{T})]}
\middle|
s_0=s
\right].
\end{equation}
The indicator term ensures that rewards are accumulated only until the goal \(\gr\) is reached or a terminal state is reached. Under this definition, the objective of reaching all the goals of a  sequence \(\gseq\) can be expressed as the sum of the individual goal values:
\begin{equation}
\label{eq:vsum_to_vtransit}
\vsum(s,\gseq,\gpi) = 
\sum\limits_{\gr\in \gseq}\ourv(s,\gr,\gpi). 
\end{equation} 
We provide a proof of ~\eqref{eq:vsum_to_vtransit} in Appendix~\ref{sec:math_proof}.


As $\ourv$ is only conditioned on a fixed-size input \((s,\gr,\gpi)\), we can use a fixed-size neural network architecture reducing the space over which the agent is required to generalize. Crucially, this decomposition does not introduce additional learning complexity. $\ourv$ corresponds to an expected discounted return with terminal conditions. As a consequence, $\ourv$ can be learned using standard temporal-difference learning methods. We name this value function  \ourcritic (\ourcriticABR).

\subsection{\ourmethod}

Our method, \ourmethod, leverages \ourcriticABR to optimize the policy not only for the final goal but also for a sequence of intermediate goals leading to it.
This is particularly beneficial during the early stages of training, when the critic is expected to provide more accurate estimates for goals that are closer to the current state. More generally, the reliability of the critic decreases for goals that are farther from the current state, which can create difficulties even at later stages of training. From the action-gap \citep{actiongap} or the signal-to-noise \citep{hiql} perspective, optimization becomes challenging when value differences are small relative to estimator noise, as happens for discounted rewards far in the future. By including intermediate goals, the actor receives more immediate learning signals that mitigate this difficulty.

For a given state-goal pair \((s, \gpi)\), we rely on a planner to compute a sequence of intermediate goals $\gseq = \text{plan}(s, \gpi)$, which is used during training only. The policy is then optimized to maximize the sum of critic values associated with each goal in the sequence:
\begin{equation}
\label{eq:actorloss}
\theta=\arg\max_{\theta} \mathbb{E}_{ a\sim \pi_\theta(.|s,\gpi)}\left[\sum_{\gr\in  \gseq = plan(s,\gpi)} \ourq^{\pi_\theta}(s,\gr,\gpi,a)\right].
\end{equation}

By linearity of expectation, the inner sum of Q-values is a sum of expectations, each defining the state-value function $\ourv^{\pi_\theta}(s,\gr,\gpi) = \mathbb{E}_{a\sim\pi_\theta}[\ourq^{\pi_\theta}(s,\gr,\gpi,a)]$. This yields the equivalent objective:
\begin{equation*}
\theta = \arg\max_{\theta} \sum_{\gr\in\gseq} \ourv^{\pi_\theta}(s,\gr,\gpi).
\end{equation*}
As shown in \eqref{eq:vsum_to_vtransit}, this objective is identical to the value function defined under the \sgmdp.

\paragraph{Hindsight Relabeling}
\label{sec:ourrelabeling}
As is common in sparse-reward GCRL, we employ hindsight experience replay to densify training signals. Transitions of the form $(s_t,a_t,\gpi,s_{t+1})$ are relabeled as $(s_t,a_t,\phi(s_{t+k}),s_{t+1})$ where $s_{t+k}$ with $k>0$ is a future state of the trajectory.  

In the case of \ourcriticABR, training requires specifying not only the goal $\gpi$ conditioning the policy but also $\gr$. 
Since the sole purpose of \ourcritic parametrized by $\psi$ is to guide the policy, we design the relabeling strategy to ensure that all critic values appearing in the actor loss are accurately learned. To this end, we train $\ourq^\psi$ on all intermediate goals along the planned sequence from the current state to the target goal, as these are precisely the values appearing in the actor objective. In addition, we also train $\ourq^\psi$ on the currently achieved goal $\phi(s)$ to densify rewards. Formally, for each sample $(s,a,s',\gpi)$ of a mini batch, we define the loss as
\begin{equation}
\label{eq:Qrelabeling}
l(s,a,s',\gpi,\psi) = 
    \sum_{
    g_r\in plan(s,\gpi) \cup \{\phi(s)\}
    }
    \left\|
    \ourq^{\psi}(s,g_r,g_\pi,a)-y_{\theta,\psi}(s',g_r,\gpi) 
    \right\|^2,
\end{equation}
where the target is given by
$$y_{\theta,\psi}(s',g_r,\gpi) =  R_{gc}(s',\gr)+\gamma\mathbbm{1}{[s'\notin \mathcal{T} \cup S_{\gr}]}\ourq^{\psi}(s',g_r,g_\pi,\pi_\theta(.|s',g_\pi)).$$
\ourmethod uses SAC to learn a \ourcritic by estimating the value defined in \eqref{eq:vtransit}, applies the relabeling strategy \eqref{eq:Qrelabeling}, and trains an actor by maximizing the objective in \eqref{eq:actorloss}.

\section{Experiments}

We empirically evaluate \ourmethodABR and several baselines in goal-conditioned environments. We test whether each method can reach near and far goals, handle goal-chaining, and avoid terminal states that lead to failure. To learn general policies, agents are trained by sampling goals uniformly across the entire goal space. 
Since testing all possible goal and start combinations is too costly, we evaluate on a fixed set of goals with increasing difficulty.

\subsection{Baselines}

We evaluate \ourmethodABR against the following baselines. 

\paragraph{SAC+HER:} 
An agent that combines Soft Actor-Critic (SAC) \citep{haarnoja_soft_2018} with Hindsight Experience Replay (HER) \citep{Andrychowicz_her}. The agent is trained to reach the final goal using off-policy actor-critic learning with goal relabeling, without exploiting any planning information. All other baselines retain the same SAC+HER backbone but introduce auxiliary objectives that leverage intermediate goals.

\paragraph{RS:}  
Unlike the SAC+HER baseline, RS leverages the planner by incorporating a Reward Shaping (RS) bonus. Reward shaping is a well-established technique that provides additional rewards to guide learning  \citep{Mataric_reward_func}, and we tailor it here specifically for sequences of goal-following tasks. Intuitively, the agent is rewarded for progressing toward the final goal and penalized for moving away from it. Progress is formally measured using the plan cost, defined as the number of intermediate goals. Let $plan(s,g)$ denote the sequence of subgoals generated by the planner from state $s$ to the final goal $g$, and define the plan cost as the number of subgoals in this sequence, $C(s,g) = |plan(s,g)|$. The shaping reward is then given by
\begin{equation}
\label{eq:rs_rew}
R_{\text{rs}}(s,a,s',g) = R_{gc}(s,a,g) + C(s,g) - C(s',g).
\end{equation}
In practice, this means that the agent receives a reward bonus of $+N$ if the plan length decreases by $N$ and $-N$ if it increases by $N$.

\paragraph{PBRS:}
This baseline incorporates Potential-Based Reward Shaping (PBRS) \citep{ng_policy_inv}, which provides a theoretical guarantee that the auxiliary shaping signal does not alter the optimal policy of the original task. To achieve this, a potential function $\Phi(s,g)$ is defined over states. Here, we set the potential as minus the remaining plan cost, $\Phi(s,g) = -C(s,g)$. Intuitively, states closer to the final goal (with a smaller plan cost) have higher potential. The augmented reward function is then defined as
\begin{equation}
\label{eq:pbrs_rew}
\begin{aligned}
R_{pbrs}(s,a,s',g) &= R_{gc}(s,a,g)+ \gamma \Phi(s',g)-\Phi(s,g) \\
                  &= R_{gc}(s,a,g)+
                   \begin{cases}
                       C(s,g) &\text{if }s' \text{ is terminal,}\\
                       C(s,g) - \gamma C(s',g) &\text{otherwise.}\\
                  \end{cases} 
\end{aligned}
\end{equation}
Compared to the RS bonus defined in \eqref{eq:rs_rew}, the shaping term in \eqref{eq:pbrs_rew} differs in two aspects required for policy invariance.
First, the immediate reward signal for progress is $C(s,g) - \gamma C(s',g)$ rather than the undiscounted difference $C(s,g) - C(s',g)$ used by RS. Second, PBRS requires that the potential of terminal states is zero, $\Phi(s,g)=0$ for all $s \in \mathcal{T}$ \citep{reward_shaping_episodic}. This ensures the total shaping contribution over any complete trajectory sums to zero, preserving the optimal policy of the original task, which is not guaranteed by the RS baseline.

\paragraph{RIS:}
This baseline incorporates ideas from "Reinforcement learning with Imagined Subgoals" (RIS) \citep{chane-sane_goal-conditioned_2021}. The core concept of the learning process is to guide the policy towards a final goal $g$ by constraining it to stay close to the behavior needed to reach an easier, intermediate goal $g_{\text{mid}}$. This is achieved by adding a Kullback-Leibler (KL) divergence penalty to the standard actor update. The policy improvement step is defined as
\[
\pi_{\theta_{k+1}} = \arg \max_{\theta} \mathbb{E}_{(s,g) \sim D, a \sim \pi_{\theta}} \left[ Q^{\pi}(s,a,g) - \alpha D_{\text{KL}}\big( \pi_{\theta}(\cdot|s,g) \ \big\| \ \pi^{\text{prior}}_k(\cdot|s,g_{\text{mid}}) \big) \right]
.\]
Here, $\pi^{\text{prior}}(\cdot|s,g_{\text{mid}})$ represents the distribution of actions that would lead from the current state $s$ to the imagined intermediate goal $g_{\text{mid}}$. Unlike the original RIS algorithm which learns a high-level policy to predict $g_{\text{mid}}$, our implementation uses the same expert planner as the other baselines. For a given state $s$ and a final goal $g$, we query the planner for the full subgoal sequence and set the imagined goal $g_{\text{mid}}$ to be the middle goal in that sequence.

\subsection{Environments}
\label{sec:envs}

We evaluate on five continuous control tasks. \dubins and \pointmaze are 2D navigation tasks where a car or a ball must reach a goal location. \antmazeu requires a quadruped to navigate to a goal location in a maze. \hopper and \walker are locomotion tasks in which the agent must reach a target $x$-position, posing a goal-chaining challenge as the optimal behaviors for reaching near and far goals differ. In all environments, a handcrafted planner provides intermediate goals. See Appendix~\ref{sec:env_details} for complete details.

\subsection{Empirical evaluation}

\begin{figure}[t]
\includegraphics[width=0.95\textwidth]{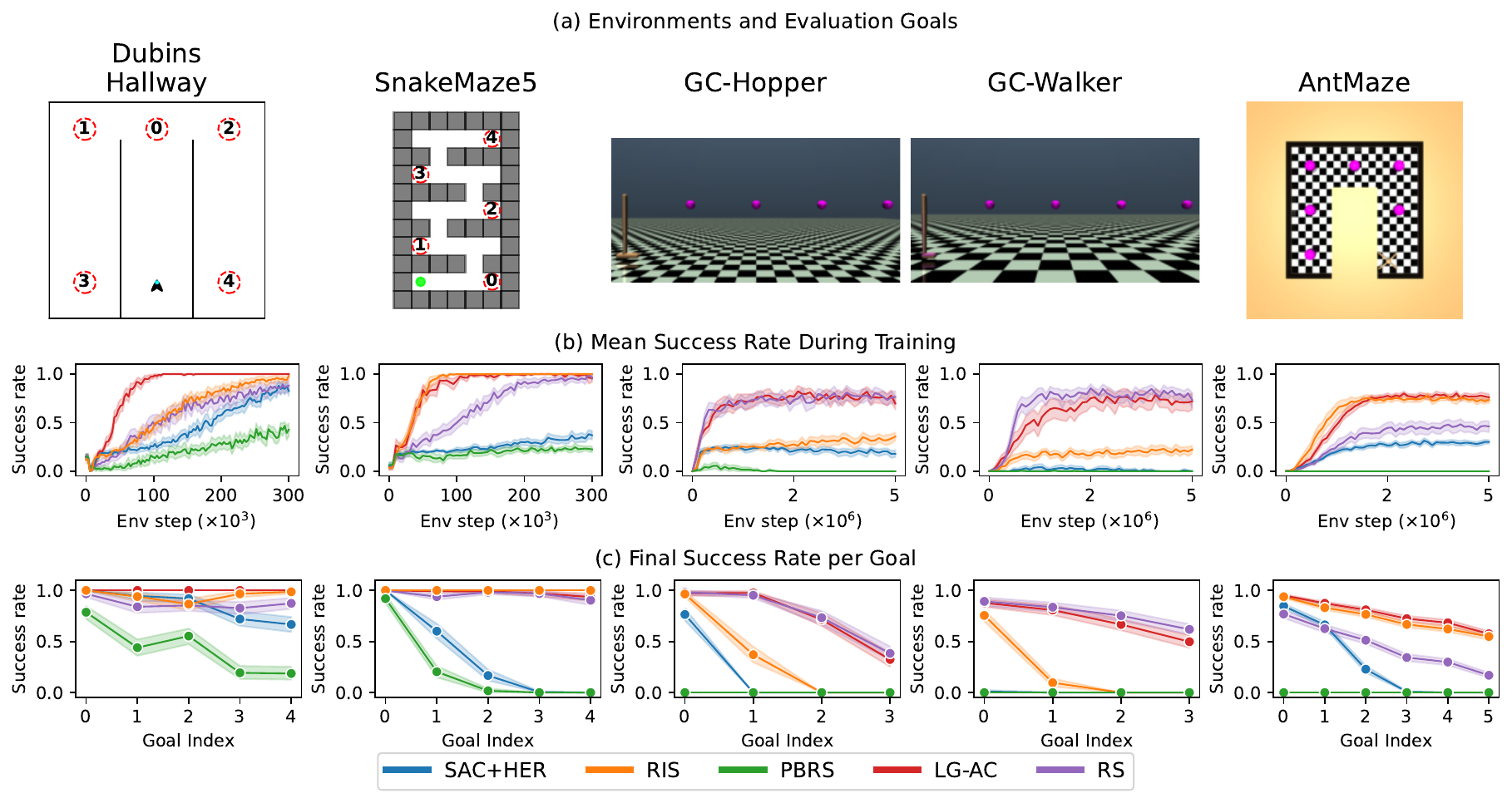}
    \caption{(a) The five goal-conditioned environments and the corresponding evaluation goal sets. (b) Mean success rate on the evaluation goal set during training. Each evaluation is repeated 10 times, with results reported as the mean and 95\% confidence interval (computed with bootstrapping) over 30 seeds. (c) Final performance per individual evaluation goal for each method, averaged over the last five evaluations. 
    In \hopper and \walker, \rs achieves the best performance with \ourmethodABR following closely behind, whereas in \antmazeu, \ris and \ourmethodABR perform best. In \dubins, \pbrs has the worst performance, often failing to reach far-away goals.}
    \label{fig:comparisonbaselines}
\end{figure}

\begin{figure}[b]
\includegraphics[width=0.85\textwidth]{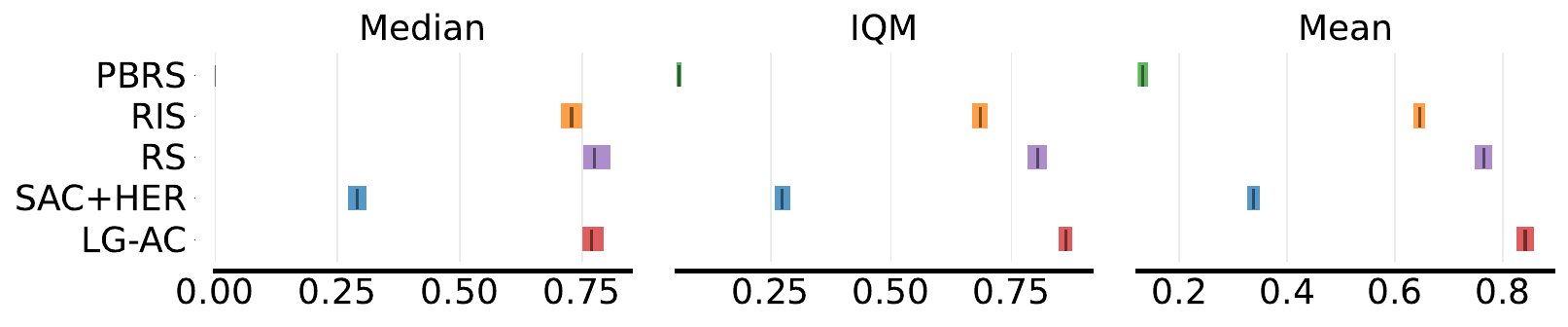}
    \caption{
    Aggregated median, interquartile mean (IQM), and mean performance across environments, computed from the last five final evaluations of each run shown in Figure~\ref{fig:comparisonbaselines}. 
    Confidence intervals are estimated using percentile bootstrap with stratified sampling \citep{agarwal_precipice} from 30 seeds. While \rs, \ris, and \ourmethodABR achieve similar median performance, \ourmethodABR outperforms the baselines on all other metrics.}
    \label{fig:aggregates}
\end{figure}

\begin{figure}[hbt]
\includegraphics[width=0.80\textwidth]{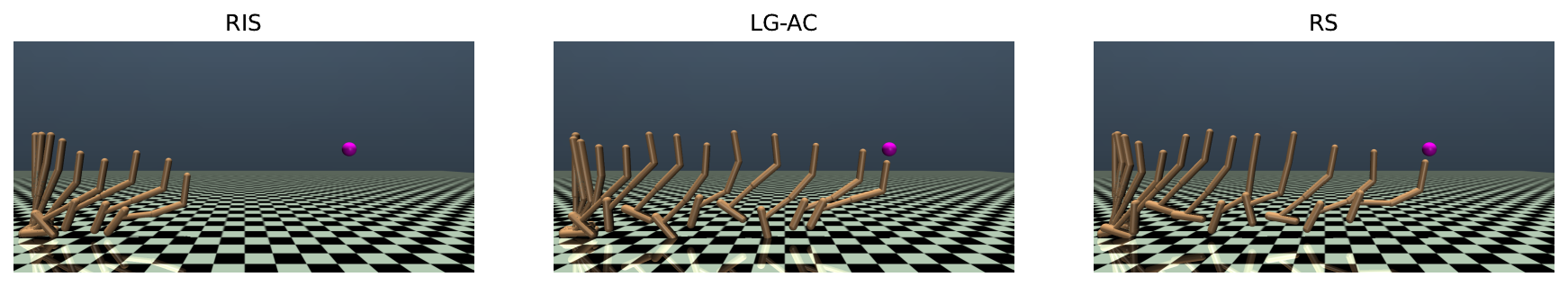}
    \caption{
    \hopper trajectories for \ris, \ourmethodABR, and \rs agents after training. \ris fails due to a myopic jump, whereas \ourmethodABR and \rs complete the task successfully.}
    \label{fig:hopper_traj}
\end{figure}

We evaluate our method on the environments defined in Section~\ref{sec:envs}. Performance is measured by the success rate on individual goals of increasing difficulty and the mean success rate during training.

\sacher performs reliably only when the target goal is temporally close to the starting configuration. For example, in  \dubins the farthest goal can be reached in approximately 40 time steps. As shown in \figurename~\ref{fig:comparisonbaselines}, \sacher achieves a high success rate across all goals in this setting. 
In the other environments, the farthest goals are more than 100 time steps away, and \sacher fails to reach these distant goals, despite often succeeding on nearby ones.

In contrast, \pbrs performs poorly across most environments, which we attribute to the formulation of the shaping bonus. First, the inclusion of the discount factor $\gamma$ creates a persistent positive reward even when the agent does not progress. If the plan length $C(s, g)$ remains unchanged between a state $s$ and its successor $s'$, the immediate shaping reward bonus $C(s, g) - \gamma C(s', g)$ simplifies to $C(s, g) (1 - \gamma)$. This persistent positive signal appears sufficient to perturb the agent. In our experiments, \pbrs underperforms relative to SAC, while the \rs baseline, which omits the $\gamma$ factor, outperforms both. This is consistent with previous findings that removing the discount factor from the PB reward can improve empirical performance \citep{PBRS_Humanoid_Locomotion}. 
A second issue arises from terminal conditions in \antmazeu, \hopper, and \walker, where falling triggers a terminal state. To maintain theoretical policy invariance, the potential of a terminal state must be zero. Consequently, upon termination, the agent receives a reward equivalent to $C(s, g)$, the total number of remaining goals in the plan. This creates a deceptive reward signal that rewards early failure. This signal severely hinders the learning process and leads to the catastrophic failures observed for \pbrs in these locomotion tasks. By addressing these issues, the RS baseline achieves success rates near 100\% in \pointmaze and \dubins. It also reaches the highest success rates in \hopper and \walker, closely followed by \ourmethodABR. In \antmazeu, while \rs outperforms \sac, it achieves lower success rates than both \ris and \ourmethodABR.

\ris performs well in \dubins, \pointmaze, and \antmazeu, but its performance drops significantly in \hopper and \walker. This failure is likely due to the main assumption in \ris  when applied to low-dimensional goals, that the best action to reach an intermediate goal is also the correct action for reaching a far-away goal. In locomotion tasks like \hopper, the best action to reach a goal in the middle of the trajectory may lead to myopic behavior, as illustrated in \figurename~\ref{fig:hopper_traj}: the agent jumps toward the final goal and subsequently fails to recover balance. In these cases, the \ris auxiliary loss provides deceptive guidance.

The \ourmethodABR method achieves success rates comparable to the best-performing method in each environment and avoids the failures seen in all other baselines. As shown in Figure~\ref{fig:aggregates}, aggregate metrics over all environments show that while \ris, \rs, and \ourmethodABR yield similar median performance, \ourmethodABR demonstrates statistically significant improvements in the mean, interquartile mean (IQM), and optimality gap as indicated by the non-overlapping confidence intervals.

\section{Conclusion}
This study investigated the problem of learning general goal-conditioned policies through formulations that distill planning guidance into a policy conditioned only on the final goal, enabling execution without a planner at deployment time.

We proposed \ourmethodABR, an asymmetric actor-critic method in which the actor is rewarded for completing a plan through a decomposition of the critic per-goal. With this formulation, the actor receives local guidance provided by the critic for each individual goal. We evaluated this approach on five goal-conditioned benchmark environments with varying levels of goal-chaining difficulty and compared it to \ris, \pbrs, and \rs.

Our empirical results indicate that \ris performs efficiently in most environments, but exhibits failure cases in \hopper and \walker, where the optimal action for reaching an intermediate goal may differ from valid actions for reaching the final goal. We also empirically confirm theoretical limitations previously identified for \pbrs. The proposed \rs variant, based on \pbrs and combined with HER, achieves strong results across environments, with the exception of \antmazeu. Overall, our method achieves the best average IQM and mean performance across the tested environments.

A limitation of our approach is the increased computational cost, which scales with the number of intermediate goals used to train both the actor and the critic.

Several directions remain for future work. 
Since our formulation allows the actor to optimize each future goal independently, one could prioritize learning on goals where the critic's predictions are trustworthy. Critic outputs may be unreliable because they correspond to insufficiently explored goals, or because the reward signal is too temporally distant to provide accurate learning feedback. We could therefore weight the critic's contributions for each future goal according to a reliability measure, for example using ensemble disagreement. Furthermore, while the use of an expert planner enables a controlled evaluation in which successes and failures can be clearly attributed to each low-level method, it would be of interest to study how \ourmethodABR behaves when the planner is learned jointly with the policy, and how robust it is to imperfect or evolving plans during training.

\bibliography{main}
\bibliographystyle{plainnat}






\appendix

\section{Proof of Value Decomposition}
\label{sec:math_proof}
In this section, we provide a proof for  Equation~\eqref{eq:vsum_to_vtransit}, which relates the value function $\vsum$ to a sum of per-goal contributions. Similarly to \cite{juozapaitis_expl_rew_dec}, we apply decomposition across rewards for intermediate goals, but also with different terminal conditions, to ensure each intermediate goal only participates to rewards only until reached. 
First, we establish a lemma that relates $g\in \gseq_t$ to a product of indicator functions encoding that $g$ has not been reached previously. This product can then be used to stop rewards from accumulating in a GC value objective once $g$ is achieved. 
\greybox{
    \paragraph{Lemma 1:}
    For any \( g \in \gseq_0 \) and any \( t \ge 0 \),
    \begin{equation*}
    \label{eq:lemma}
    \mathbbm{1}[g \in \gseq_t]
    =
    \prod_{i=1}^{t} \mathbbm{1}[s_i \notin S_g],
    \quad
    \text{where the empty product $(t=0)$ is defined as 1.}
    \end{equation*}
    \paragraph{Proof:} By induction on $t$. For $t=0$, $g\in \gseq_0$, so the left side is 1 and the right is the empty product, which is 1. Assume the statement holds for $t-1$. From the update rule  (Eq.~\ref{eq:goal_transition}),$$g\in\gseq_t \iff (g\in\gseq_{t-1})\land (s_t\notin S_g).$$
    Using the induction hypothesis, 
    $$\mathbbm{1}[g\in\gseq_t] = \mathbbm{1}[g\in\gseq_{t-1}] \cdot \mathbbm{1}[s_t\notin S_g] = \Big(\prod_{i=1}^{t-1}\mathbbm{1}[s_i\notin S_g]\Big) \cdot \mathbbm{1}[s_t\notin S_g] = \prod_{i=1}^{t}\mathbbm{1}[s_i\notin S_g]. \quad \square$$
    }
With this lemma, we now derive the desired decomposition.
\begin{align*}
\vsum(s,\gseq,\gpi)&=
\operatorname*{\mathbb{E}}_{\substack{
a_t\sim\pi(\cdot\mid s_t,\gpi) \\
}}
\Big[
    \sum_{t=0}^{\infty} \gamma^t\rsum(s_{t+1},\gseq_t) 
    \prod_{i=1}^t \mathbbm{1}[s_i\notin \mathcal{T}]
    \mid  \substack{s_0=s\\\gseq_0=\gseq}{}
\Big]
\\
&=\operatorname*{\mathbb{E}}_{\substack{
a_t\sim\pi(\cdot\mid s_t,\gpi) \\
}}
\Big[
    \sum_{t=0}^{\infty}  
    \gamma^t
    \Big(
        \sum\limits_{\gr\in \gseq_t}
        \rgc(s_{t+1},\gr)
    \Big)
    \prod_{i=1}^{t} \mathbbm{1}[s_i\notin \mathcal{T}]
    \mid  \substack{s_0=s\\\gseq_0=\gseq}{}
\Big]
\\
&
\text{According to \eqref{eq:goal_transition} }, \forall t\geq0, \gseq_t\subset \gseq_0 = \gseq
\\
&=\operatorname*{\mathbb{E}}_{\substack{
a_t\sim\pi(\cdot\mid s_t,\gpi) \\
}}
\Big[
    \sum_{t=0}^{\infty}  
    \gamma^t
    \Big(
        \sum\limits_{\gr\in \gseq}
        \mathbbm{1}[\gr \in \gseq_t]
        \rgc(s_{t+1},\gr)
    \Big)
    \prod_{i=1}^{t} \mathbbm{1}[s_i\notin \mathcal{T}]
    \mid  \substack{s_0=s\\\gseq_0=\gseq}{}
\Big]
\\
&=
\sum\limits_{\gr\in \gseq}
\operatorname*{\mathbb{E}}_{\substack{
a_t\sim\pi(\cdot\mid s_t,\gpi) \\
}}
\Big[
    \sum_{t=0}^{\infty}  
    \gamma^t
    \Big(
        \mathbbm{1}[\gr \in \gseq_t]
        \rgc(s_{t+1},\gr)
    \Big)
    \prod_{i=1}^{t} \mathbbm{1}[s_i\notin \mathcal{T}]
    \mid  \substack{s_0=s\\\gseq_0=\gseq}{}
\Big].
\\
& \text{Using lemma  1,}
\\
 &=
\sum\limits_{\gr\in \gseq}
\operatorname*{\mathbb{E}}_{\substack{
a_t\sim\pi(\cdot\mid s_t,\gpi)
}}
\Big[
    \sum_{t=0}^{\infty}  
    \gamma^t
    \Big(
        \rgc(s_{t+1},\gr)
        \prod_{i=1}^{t}\mathbbm{1}[s_i\notin S_{\gr}]
    \Big)
    \prod_{i=1}^{t} \mathbbm{1}[s_i\notin \mathcal{T}]
    \mid  s_0=s
\Big]
\\
&=
\sum\limits_{\gr\in \gseq}
\operatorname*{\mathbb{E}}_{\substack{
a_t\sim\pi(\cdot\mid s_t,\gpi)
}}
\Big[
    \sum_{t=0}^{\infty}  
    \gamma^t
    \Big(
        \rgc(s_{t+1},\gr)
        \prod_{i=1}^{t}\mathbbm{1}[s_i\notin (S_{\gr}\cup\mathcal{T})]
    \Big)
    \mid  s_0=s
\Big].
\\
\end{align*}
By defining the \ourcritic,
\[
\ourv(s,\gr,\gpi)
\triangleq
\mathbb{E}_{a_t\sim\pi(\cdot\mid s_t,\gpi)}
\left[
\sum_{t=0}^{\infty}
\gamma^t
\rgc(s_{t+1},\gr)
\prod_{i=1}^{t}
\mathbbm{1}[s_i\notin (S_{\gr}\cup\mathcal{T})]
\middle|
s_0=s
\right].
\]
We therefore obtain the decomposition
\[
\vsum(s,\gseq,\gpi) = 
\sum\limits_{\gr\in \gseq}
\ourv(s,\gr,\gpi). \quad \square
\]
This decomposition shows that the value function for a set of goals is the sum of independent per‑goal value functions, each measuring the expected discounted reward from that goal until it is reached or a terminal state is encountered.

\section{Relabeling Strategies}

\begin{figure}[htbp]
\includegraphics[width=0.95\textwidth]{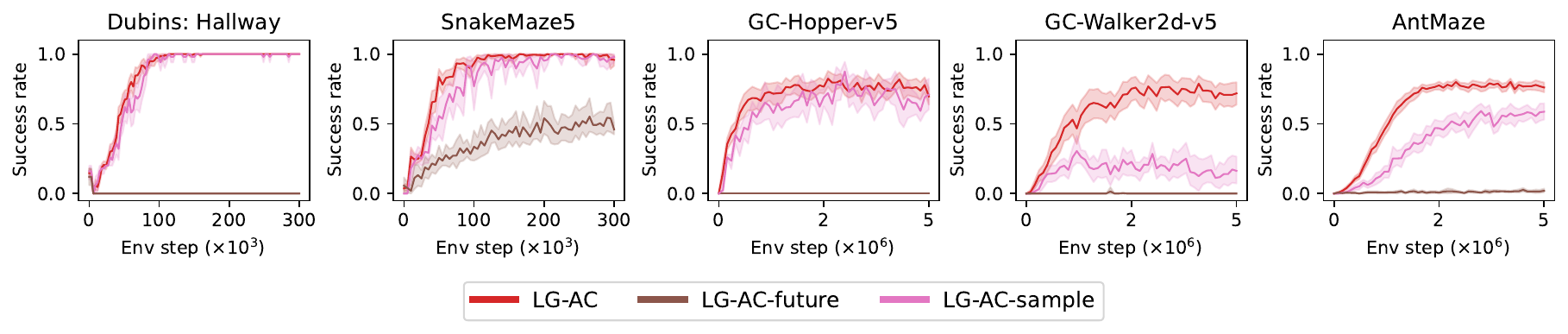}
    \caption{
     Mean success rate on the evaluation goal set during training. Each evaluation was repeated 10 times, with results reported as the mean and 95\% confidence interval (computed with bootstrapping) over 10 seeds.
     \asymfuture leads to low SR, only learning in \pointmaze. \ourmethodABR performs equally or better compared to \asymsample.}
    \label{fig:her_study}
\end{figure}

This section investigates the selection of relabeling techniques for the \ourcritic, denoted as $Q(s,a,\gr,\gpi)$. For $\gpi$, the final goal of the episode, we propose to always sample using HER. For $\gr$, the goal conditioning the reward, we investigate three distinct approaches.

The first strategy, \asymfuture, randomly selects a goal achieved later in the same trajectory. Another variant, \asymsample, samples $\gr$ from $plan(s,\gpi)$ so that the critic is trained on the values optimized by the actor. Finally, the \ourmethodABR strategy trains the critic on all goals within $plan(s,\gpi)$, as described in Section~\ref{sec:ourrelabeling}, to reduce the risk of encountering out-of-distribution values during actor optimization.

As shown in \figurename~\ref{fig:her_study}, \asymfuture achieves the lowest performance, suggesting that sampling goals from the current plan is critical. Moreover,  \ourmethodABR consistently performs similarly to or better than \asymsample, but at a higher computational cost.

\section{Environment details}
\label{sec:env_details}

\begin{figure}[h]
\includegraphics[width=0.95\textwidth]{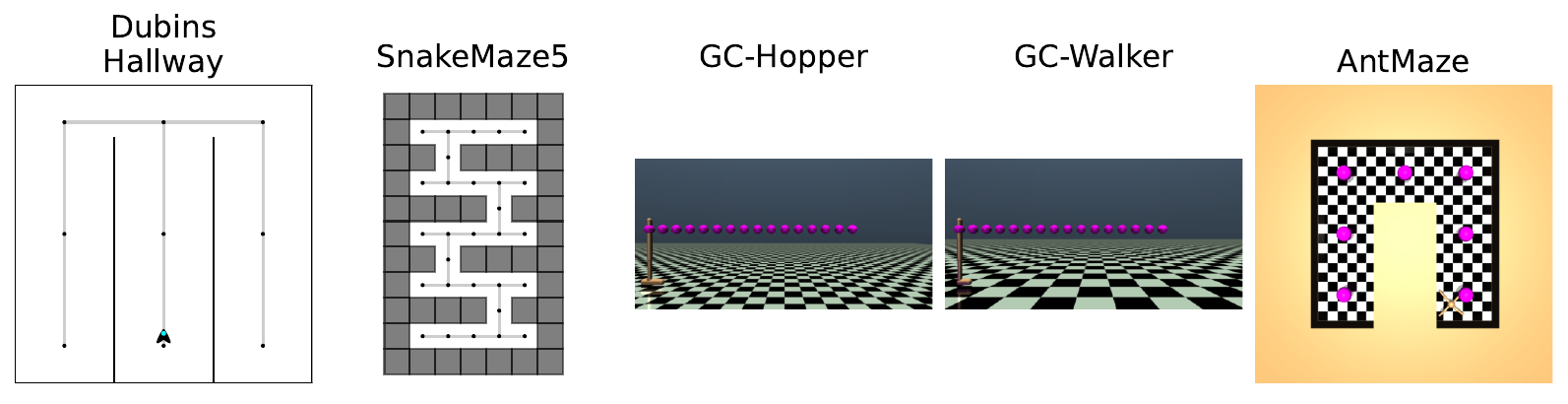}
    \caption{Handcrafted graphs used for planning in \dubins, \pointmaze, and \antmazeu. For \hopper and \walker, the planner outputs uniformly spaced points between the agent and the final goal.}
    \label{fig:planning}
\end{figure}

\paragraph{\dubins:}  A navigation task in which the agent controls a car in a 2D maze. The state $s=\{x,y,\cos(\theta),\sin(\theta)\}$ includes the position and orientation of the agent. Each goal $g$ is defined as a position $(x,y)$ and is considered reached if the agent is within a ball of radius $0.1$ centered at that position. 
At each step, the agent advances at a fixed speed, and the action specifies the rate of change of its orientation $\dot{\theta}$.
During a {training} episode, both the initial position of the agent and the goal are randomly sampled within the maze boundaries. During an {evaluation} episode, the agent is tested on a fixed start state and multiple goal locations illustrated in \figurename~\ref{fig:comparisonbaselines}. If the agent hits a wall, it stays stuck until the episode ends. The state is only terminal when the agent reaches the goal.

\paragraph{\pointmaze:} The agent controls a ball navigating in a 2D maze \citep{gymnasium_robotics2023github}. The state $s=\{x,y,\dot{x},\dot{y}\}$ includes the position and velocity of the agent. Each goal $g$ is defined as an $(x,y)$ position and is considered reached if the agent is within a ball of radius $0.45$ centered at that position. The action represents the linear force exerted on the ball in the x and y directions. 
In \cite{gymnasium_robotics2023github}, the agent's velocity was limited to the range $[-5,5]$ m/s. Within this range, the agent could easily take sharp turns at maximum speed, making the task relatively simple. To increase the difficulty, we extended this limit to $[-10, 10]$ m/s\footnote{The upper bound was determined empirically; beyond this value, collisions become unstable and the agent can occasionally pass through maze walls.}, requiring the agent to anticipate turns and decelerate in advance. 
During a {training} episode, both the agent's initial position and the goal are randomly sampled within the maze boundaries. During an {evaluation} episode, the agent is tested on multiple goal locations illustrated in \figurename~\ref{fig:comparisonbaselines}. The state is only terminal when the agent reaches the goal.

\paragraph{\hopper:} The agent controls a two-dimensional, one-legged figure based on \cite{gymnasium_robotics2023github}. The state consists of the positions and velocities of the robot’s body parts. 
The goal is defined as the difference between the agent’s current $x$-location and a desired location $x_{g}$ that remains fixed during an episode; notably, the absolute $x$-position is excluded from the state to promote generalization. Actions represent the torques applied to the hinge joints. An episode terminates either when the goal is reached, defined as $(x-x_{g})<0.1$, or when the hopper’s posture becomes unhealthy: the height of the hopper falls below $0.7$ or the torso angle leaves the range $[-0.2,0.2]$. During training, a random goal is sampled in the range $[0,6.5]$. During evaluation, the agent is tested on the specific goals $[1.5,3,4.5,6]$. Hopper poses a goal chaining challenge because the optimal behavior needed to reach nearby goals differs from that needed to reach far ones.

\paragraph{\walker:} The agent controls a two-legged version of the Hopper. The state, actions, and goals follow the same setup as \hopper, including the training and evaluation procedures.

\paragraph{\antmazeu:} The agent controls the ant quadruped from \cite{towers2024gymnasium}, in a maze, following the implementation in \cite{gymnasium_robotics2023github}. The state consists of the positions and velocities of the robot’s body parts. The goal corresponds to an $(x,y)$ location. Actions represent the torques applied to the hinge joints. An episode terminates either when the goal is reached, defined as $(x-x_{g})<0.5$, or when the robot’s posture becomes unhealthy: the height of the ant is not in the closed interval $[0.2,1]$. During training, goals are sampled by adding uniform noise in $[-0.25,0.25]$ to the center of a random grid cell. During evaluation, the agent is tested on the goals shown in Figure~\ref{fig:comparisonbaselines}.

\paragraph{Handcrafted Planner} In \dubins, \pointmaze, and \antmazeu, the planner computes the shortest path using the expert graph (see Figure~\ref{fig:planning}). In \hopper and \walker, the planner instead sets intermediate goals evenly spaced between the agent and the final goal.

\section{Hyperparameters}
\label{sec:hyperparameters}
\begin{table}[htbp]
\centering
\caption{Common hyperparameters. Exceptions are noted in the rightmost column.}
\label{tab:hp_common}
\begin{tabular}{lll}
\toprule
Hyperparameter           & Common value          & Notes / Exceptions \\
\midrule
Random actions           & 50k                  & 5k for \dubins and \pointmaze \\
Critic hidden size       & 256                   & --- \\
Policy hidden size       & 256                   & --- \\
Activation function      & ReLU                  & --- \\
Discount factor ($\gamma$) & 0.99                & --- \\
Critic learning rate     & \(3\times10^{-4}\)    & --- \\
Policy learning rate     & \(3\times10^{-4}\)    & --- \\
Temperature learning rate & \(3\times10^{-4}\)   & --- \\
\ris-specific $\alpha$   & \(2^{-8}\)            &(only for \ris) \\
HER relabel strategy     & 20\% original, 80\% future         & \antmazeu: 20\% original, 40\% future, 40\% random \\
\bottomrule
\end{tabular}
\end{table}

\begin{table}[htbp]
\centering
\caption{Batch sizes per environment and method. \ourmethodABR uses a fixed batch size of 256 for all environments; the other methods have tuned values.}
\label{tab:hp_batch}
\begin{tabular}{lccccc}
\toprule
Environment & \ourmethodABR & \ris & \rs & \pbrs & \sacher \\
\midrule
\dubins     & 256 & 512 & 256 & 256 & 512 \\
\pointmaze  & 256 & 256 & 256 & 256 & 256 \\
\hopper     & 256 & 512 & 512 & 256 & 256 \\
\walker     & 256 & 256 & 256 & 256 & 256 \\
\antmazeu   & 256 & 256 & 256 & 256 & 256 \\
\bottomrule
\end{tabular}
\end{table}

The hyper‑parameters used for each environment are presented in Table~\ref{tab:hp_common} and Table~\ref{tab:hp_batch}. All baselines and our method share the same hyper‑parameters, taken from the original SAC paper \citep{haarnoja_soft_2018}, with the following exceptions. The number of random actions taken before learning is scaled to the complexity of the environment. For most environments, we use future HER \citep{Andrychowicz_her}. For \antmazeu, we employ the same relabeling strategy as in \ris \citep{chane-sane_goal-conditioned_2021}: goals are relabeled either with future states from the same trajectory or with random achieved goals from the replay buffer. We observed that this variant is particularly beneficial in \antmazeu, while it has little effect in other environments. 
In addition, the RIS method introduces an extra hyper‑parameter $\alpha$, the procedure for tuning this value is detailed in Section~\ref{sec:ris_alpha_tuning}.
Finally, batch sizes are tuned per (baseline, environment) pair, as explained in Section~\ref{sec:batch_size_on_perf}.

\section{Impact of batch size on performance}
\label{sec:batch_size_on_perf}

\begin{table}[ht]
\centering
\caption{
Analysis of the effective batch size for \ourmethodABR. The effective batch size is the total number of targets processed per update; it is calculated by multiplying the base batch size (256) by the average plan length per transition. Statistics (mean $\pm$ std) are computed across all runs shown in Figure~\ref{fig:comparisonbaselines}.
}
\label{tab:effective_batch_size}
\begin{tabular}{lcc}
\toprule
\textbf{Environment} & \textbf{Avg. Goals per Transition} & \textbf{Effective Batch Size} \\
\midrule
\dubins      & $2.44 \pm 0.02$ & $624 \pm 6$   \\
\pointmaze  & $3.96 \pm 0.11$ & $1,013 \pm 27$ \\
\hopper      & $5.14 \pm 0.25$ & $1,315 \pm 64$ \\
\walker      & $5.03 \pm 0.40$ & $1,286 \pm 102$ \\
\antmazeu       & $3.01 \pm 0.11$ & $771 \pm 27$  \\
\bottomrule
\end{tabular}
\end{table}

\begin{table*}[th]
\centering
\caption{Success rate of all baseline algorithms across all environments under varying batch sizes. Each cell reports the mean and standard deviation over 10 random seeds for a given (algorithm, environment, batch size) configuration. Bold values indicate either (i) the highest mean performance among the tested batch sizes for a given (algorithm, environment) pair, or (ii) results that are not statistically significantly different from the best-performing batch size for that pair.}
\label{tab:batchsize}
\begin{tabular}{llccc}
\toprule
\textbf{Method} & \textbf{Environment} & \textbf{Batch 256} & \textbf{Batch 512} & \textbf{Batch 1024} \\
\midrule
\sacher & \dubins & $0.80 \pm 0.16$ & $\boldsymbol{0.90} \pm 0.12$ & $0.78 \pm 0.16$ \\
 & \pointmaze & $\boldsymbol{0.38} \pm 0.09$ & $\boldsymbol{0.38} \pm 0.08$ & $0.33 \pm 0.10$ \\
 & \hopper & $\boldsymbol{0.18} \pm 0.10$ & $\boldsymbol{0.22} \pm 0.08$ & $\boldsymbol{0.19} \pm 0.10$ \\
 & \walker & $\boldsymbol{0.00} \pm 0.00$ & $\boldsymbol{0.00} \pm 0.00$ & $\boldsymbol{0.02} \pm 0.06$ \\
 & \antmazeu & $\boldsymbol{0.30} \pm 0.07$ & $\boldsymbol{0.31} \pm 0.06$ & $\boldsymbol{0.30} \pm 0.08$ \\
\midrule
\pbrs & \dubins & $\boldsymbol{0.40} \pm 0.24$ & $\boldsymbol{0.34} \pm 0.13$ & $0.28 \pm 0.19$ \\
 & \pointmaze & $\boldsymbol{0.24} \pm 0.08$ & $\boldsymbol{0.23} \pm 0.11$ & $\boldsymbol{0.21} \pm 0.09$ \\
 & \hopper & $\boldsymbol{0.00} \pm 0.00$ & $\boldsymbol{0.00} \pm 0.00$ & $\boldsymbol{0.00} \pm 0.00$ \\
 & \walker & $\boldsymbol{0.00} \pm 0.00$ & $\boldsymbol{0.00} \pm 0.00$ & $\boldsymbol{0.00} \pm 0.00$ \\
 & \antmazeu & $\boldsymbol{0.00} \pm 0.00$ & $\boldsymbol{0.00} \pm 0.00$ & $\boldsymbol{0.00} \pm 0.00$ \\
\midrule
\ris & \dubins & $0.97 \pm 0.07$ & $\boldsymbol{1.00} \pm 0.00$ & $0.98 \pm 0.07$ \\
 & \pointmaze & $\boldsymbol{1.00} \pm 0.00$ & $\boldsymbol{1.00} \pm 0.00$ & $\boldsymbol{1.00} \pm 0.00$ \\
 & \hopper & $0.30 \pm 0.13$ & $\boldsymbol{0.37} \pm 0.14$ & $\boldsymbol{0.35} \pm 0.12$ \\
 & \walker & $\boldsymbol{0.22} \pm 0.11$ & $0.17 \pm 0.13$ & $\boldsymbol{0.25} \pm 0.13$ \\
 & \antmazeu & $\boldsymbol{0.73} \pm 0.09$ & $\boldsymbol{0.77} \pm 0.07$ & $0.72 \pm 0.10$ \\
\midrule
\rs & \dubins & $\boldsymbol{0.90} \pm 0.13$ & $\boldsymbol{0.89} \pm 0.15$ & $\boldsymbol{0.90} \pm 0.14$ \\
 & \pointmaze & $\boldsymbol{0.98} \pm 0.05$ & $\boldsymbol{0.97} \pm 0.06$ & $0.95 \pm 0.07$ \\
 & \hopper & $0.69 \pm 0.19$ & $\boldsymbol{0.75} \pm 0.21$ & $\boldsymbol{0.77} \pm 0.16$ \\
 & \walker & $\boldsymbol{0.83} \pm 0.14$ & $\boldsymbol{0.83} \pm 0.16$ & $\boldsymbol{0.80} \pm 0.18$ \\
 & \antmazeu & $\boldsymbol{0.49} \pm 0.19$ & $\boldsymbol{0.42} \pm 0.26$ & $0.23 \pm 0.22$ \\
\bottomrule
\end{tabular}
\end{table*}

We analyze the impact of batch size to ensure a fair comparison between methods. As detailed in Section~\ref{sec:ourrelabeling}, the loss for each transition $(s, a, s', g_\pi)$ in \ourcriticABR incorporates all intermediate goals. This increases the number of targets processed per update, resulting in a larger effective batch size (see Table~\ref{tab:effective_batch_size}) and higher computational cost. Consequently, a direct comparison with baselines would be biased if performance gains were simply a byproduct of this increased computation. To control for this factor, we independently tuned the batch size for each
 baseline (see Table~\ref{tab:batchsize}). For each method and environment, we selected the smallest batch size that yielded performance statistically equivalent to its best-performing configuration. This protocol ensures a fair comparison while avoiding unnecessary computational overhead. 
 
\section{Impact of the KL penalty factor on RIS performance}
\label{sec:ris_alpha_tuning}

\begin{table}[htpb]
\centering
\caption{Sensitivity analysis of the \ris baseline to the KL penalty factor $\alpha$ under different reward specifications. We report the mean and standard deviation of the success rate, aggregated over two environments (\pointmaze and \dubins) with 10 independent runs each. The ideal range for $\alpha$ differs across reward specifications.
}
\label{tab:ris_sensitivity}
\renewcommand{\arraystretch}{1.25}
\begin{tabular}{l c cc}
\hline
\textbf{Method} & $\boldsymbol{\alpha}$ & \textbf{Positive Reward} & \textbf{Negative Reward} \\
\hline
\multirow{17}{*}{\ris}
 & $2^{-12}$ & $0.85 \pm 0.17$ & $0.25 \pm 0.09$ \\
  & $2^{-11}$ & $\boldsymbol{0.98 \pm 0.06}$ & $0.25 \pm 0.09$ \\
  & $2^{-10}$ & $\boldsymbol{0.98 \pm 0.07}$ & $0.26 \pm 0.10$ \\
  & $2^{-9}$ & $\boldsymbol{0.98 \pm 0.06}$ & $0.24 \pm 0.08$ \\
  & $2^{-8}$ & $\boldsymbol{0.99 \pm 0.05}$ & $0.26 \pm 0.10$ \\
  & $2^{-7}$ & $\boldsymbol{0.98 \pm 0.06}$ & $0.26 \pm 0.09$ \\
  & $2^{-6}$ & $\boldsymbol{0.97 \pm 0.09}$ & $0.36 \pm 0.16$ \\
  & $2^{-5}$ & $\boldsymbol{0.99 \pm 0.05}$ & $0.46 \pm 0.26$ \\
  & $2^{-4}$ & $0.95 \pm 0.13$ & $\boldsymbol{0.70 \pm 0.27}$ \\
  & $2^{-3}$ & $0.82 \pm 0.20$ & $\boldsymbol{0.72 \pm 0.32}$ \\
  & $2^{-2}$ & $0.47 \pm 0.22$ & $\boldsymbol{0.74 \pm 0.31}$ \\
  & $2^{-1}$ & $0.24 \pm 0.13$ & $\boldsymbol{0.76 \pm 0.29}$ \\
  & $2^{0}$ & $0.19 \pm 0.07$ & $\boldsymbol{0.74 \pm 0.29}$ \\
  & $2^{1}$ & $0.15 \pm 0.09$ & $\boldsymbol{0.69 \pm 0.33}$ \\
  & $2^{2}$ & $0.09 \pm 0.09$ & $0.60 \pm 0.40$ \\
  & $2^{3}$ & $0.09 \pm 0.09$ & $0.63 \pm 0.37$ \\
  & $2^{4}$ & $0.08 \pm 0.10$ & $0.61 \pm 0.29$ \\
\hline
\ourmethodABR & -- & $0.99 \pm 0.04$ & $0.99 \pm 0.03$ \\
\sacher & -- & $0.58 \pm 0.25$ & $0.28 \pm 0.10$ \\
\hline
\end{tabular}
\end{table}

We evaluate the sensitivity of the RIS baseline to its KL penalty factor $\alpha$ under different reward specifications. In the original work, the authors use a negative reward function $R(s',g)=\mathbbm{-1}[s'\notin g]$ and set $\alpha=1$. In our study, we primarily employ a positive reward function $R(s',g)=\mathbbm{1}[s'\in g]$. As illustrated in Table~\ref{tab:ris_sensitivity}, RIS performance varies smoothly with $\alpha$, remaining near-optimal within a specific range for both reward types. 

In the negative reward setting, we observe that $\alpha=1$ falls within the near-optimal range, which aligns with the choice made in the original article. However, the range of effective $\alpha$ values differs between the positive and negative reward settings, suggesting that RIS requires specific tuning depending on the reward type. We also note a slight decrease in RIS performance when using negative rewards. In contrast, \ourmethodABR does not introduce additional hyperparameters and maintains stable performance across both reward specifications. For a fair comparison in the main article, all experiments involving the RIS baseline use positive rewards with $\alpha=2^{-8}$.


\end{document}